\documentclass[letterpaper, 10 pt, conference]{ieeeconf}  

\IEEEoverridecommandlockouts                              

\usepackage{graphics} 
\usepackage{epsfig} 
\usepackage{mathptmx} 
\usepackage{times} 
\usepackage{amsmath} 
\usepackage{amssymb}  
\usepackage{graphicx}
\usepackage{booktabs}
\usepackage{multirow}
\usepackage{subcaption}
\usepackage{bbding}

\usepackage{marvosym}

\title{\LARGE \bf
RTPolicy: Real-Time (32.9 Hz) Multi-View Visuomotor Policy for Robotic Manipulation on Edge Devices
}

  \author{
  Xianbao Hou$^{1,3,*}$, Zehao Ni$^{2,3,*}$, Yonghao He$^{3,\dagger}$, Lingfeng Qian$^{3}$,\\
  Xin Shi$^{3}$, Hu Su$^{4}$, Wei Sui$^{3,\text{\Letter}}$, Cong Yang$^{1,\text{\Letter}}$\\[6pt]
  {$^{1}$School of Future Science and Engineering, Soochow University}\\
  {$^{2}$Shanghai Research Institute for Intelligent Autonomous Systems, Tongji University}\\
  {$^{3}$D-Robotics}\\
  {$^{4}$State Key Laboratory of Multimodal Artificial Intelligence Systems,}\\
  {\phantom{$^{4}$}Institute of Automation, Chinese Academy of Sciences}\\[4pt]
  {$^{*}$Equal contribution.\quad $^{\dagger}$Project lead.\quad $^{\text{\Letter}}$Corresponding authors.}}

\begin{document}

\maketitle
\thispagestyle{empty}
\pagestyle{empty}

\begin{abstract}
Low inference latency is critical for closed-loop robotic manipulation, especially under asynchronous inference, where action chunks must be updated promptly to correct accumulated deviations.
However, existing visuomotor policies rely on computationally expensive visual backbones, resulting in high inference latency and deployment overhead.
In this paper, we propose RTPolicy, a multi-view visuomotor policy that attains accurate manipulation with high inference efficiency.
Its core design combines view-aware feature fusion with visual compression to aggregate complementary information across camera views into a compact token representation.
The representation conditions a flow-matching action generator with the robot state and flow time, reducing the cost of action generation.
Extensive experiments demonstrate that this design reduces the computational cost of visual representation and action generation while preserving visual information required for manipulation.
On the RoboTwin~2.0 single-task clean evaluation across 50 tasks, RTPolicy attains a mean success rate of 52.26\%, outperforming the strongest leaderboard baseline $\pi_0$ (46.4\%).
We further evaluate RTPolicy under asynchronous streaming inference on a real bimanual platform, where RTPolicy attains a mean success rate of 72.0\% and sustains real-time inference at 30.35 ms (32.9 Hz) on the RDK S600 and 149.72 ms (6.7 Hz) on the RDK S100, both in INT16. Project Page: https://d-robotics-ai-lab.github.io/RTPolicy.project.

\end{abstract}

\section{Introduction}

Precise manipulation requires policies that learn from demonstrations while maintaining temporal coherence and responding to new visual observations, especially under asynchronous inference.
Visuomotor policies therefore predict short action trajectories and execute them in a receding horizon, as in action chunking~\cite{zhao2023learning}.
Closed-loop correction~\cite{ross2010efficient,ross2011reduction} keeps the executed plan consistent with the current observation.
Inference latency therefore directly determines how quickly fresh observations can affect the executed trajectory.

\begin{figure}[t]
  \centering
  \includegraphics[width=\columnwidth]{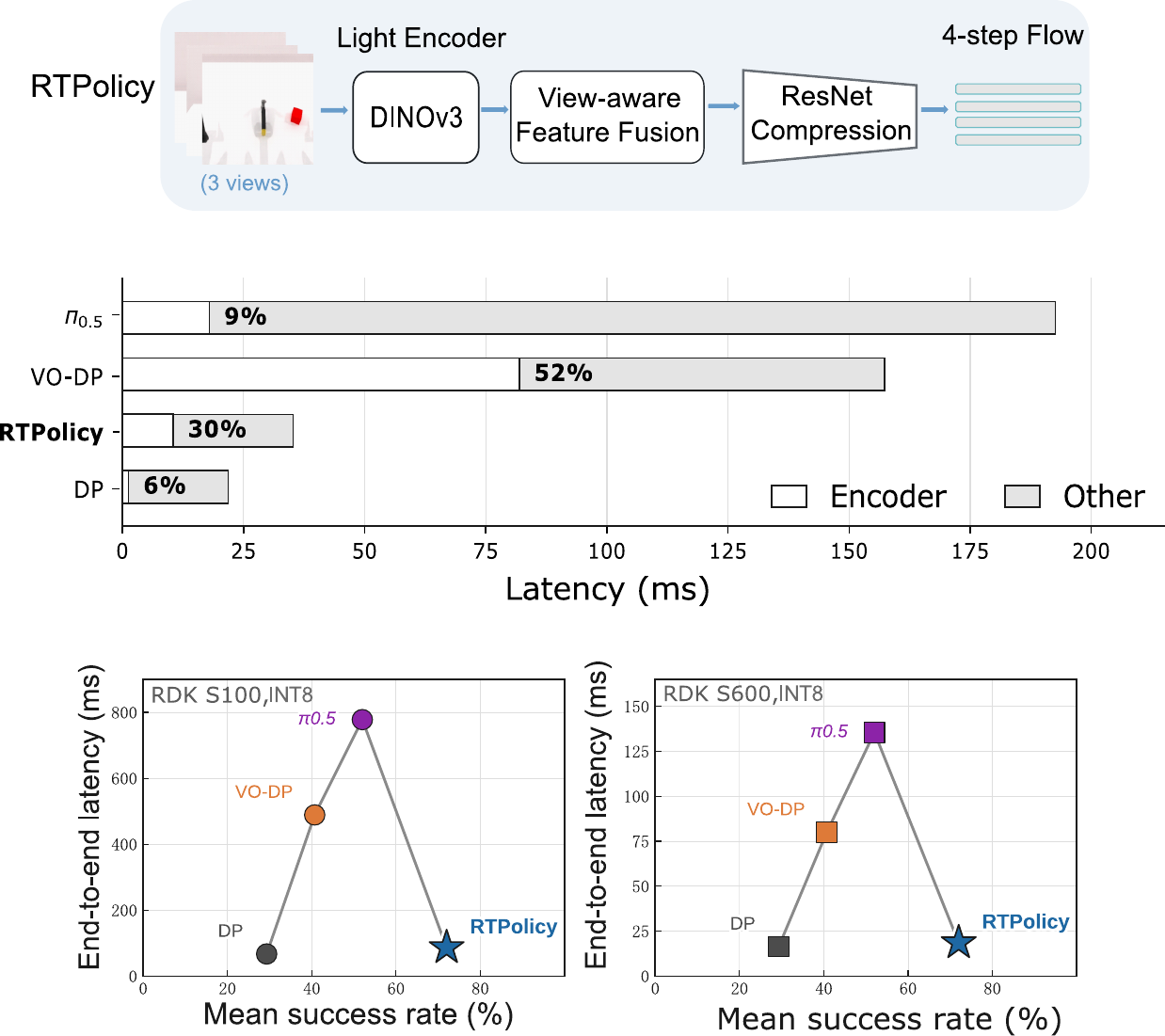}
  \caption{RTPolicy design, visual encoding cost, and deployment performance.
  The top panel shows the lightweight encoder, view-aware feature fusion, visual compression, and four-step flow generation.
  The middle panel shows that VO-DP spends 52\% of its latency on the encoder, while $\pi_{0.5}$ incurs higher latency from multi-view visual conditioning despite a smaller encoder share.
  The bottom panels report end-to-end latency against mean success rate on the RDK S100 and RDK S600 in INT8.
  RTPolicy combines the highest mean success rate with low latency on both devices.
  The success rates use the GPU evaluation configuration rather than per-device task evaluations.}
  \vspace{-2mm}
  \label{fig:teaser}
\end{figure}

Current visuomotor policies face crucial and coupled sources of inference cost (Fig.~\ref{fig:teaser}).
Large visual backbones~\cite{ni2025vodp} can make visual encoding a substantial part of latency.
Multi-view image conditioning in $\pi_{0.5}$~\cite{black2025pi05} increases the number of visual features that the action model must process.
Iterative denoising and flow sampling~\cite{wang2024onedp,zhang2024flowpolicy,su2025freqpolicy} add further computation before an action chunk can be executed.

Existing methods address individual sources of inference cost.
Consistency and flow-based policies~\cite{wang2024onedp,zhang2024flowpolicy,su2025freqpolicy} reduce the number of action generation steps, but generally leave the visual encoder and multi-view conditioning path unchanged.
Diffusion Policy (DP)~\cite{chi2023diffusion} offers low latency, but its mean success rate is 29.3\% and its single-view setup does not address the cost of multi-view conditioning (Fig.~\ref{fig:teaser}).
Achieving low latency and high manipulation accuracy with multi-view inputs remains a challenge for deployment on edge devices.

\begin{figure*}[t]
  \centering
  \includegraphics[width=\textwidth]{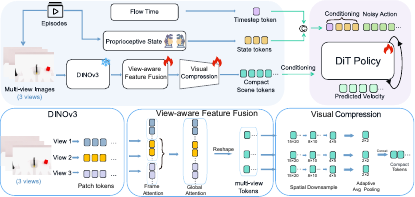}
  \caption{Overview of RTPolicy.
  A frozen DINOv3 ConvNeXt backbone extracts dense features from multi-view images.
  View-aware feature fusion refines each view and exchanges information across views before visual compression produces a compact set of visual tokens.
  A Diffusion Transformer uses these tokens, a fixed robot-state prefix, and the flow time to predict the velocity of the future action chunk.
  Four integration steps transform Gaussian noise into actions.}
  \vspace{-2mm}
  \label{fig:overview}
\end{figure*}

Hence, we propose RTPolicy, a visuomotor framework that jointly reduces the costs of visual encoding, multi-view conditioning, and iterative action generation.
It uses a frozen DINOv3 ConvNeXt encoder~\cite{simeoni2025dinov3} to obtain visual features from multi-view images.
VGGT~\cite{wang2025vggt} and VO-DP~\cite{ni2025vodp} have demonstrated the value of geometric
visual representations for perception and manipulation.
To preserve spatial information across views, our view-aware feature fusion module assigns a
learnable embedding to each view and exchanges information within and across views.
A visual compression module then compresses the fused representation into a compact visual condition before it feeds a flow-based Diffusion Transformer.
This framework reduces visual conditioning cost while retaining spatial information for precise manipulation (Fig.~\ref{fig:teaser}).
In simulation and on a real bimanual platform, RTPolicy attains the highest mean success rate among the compared methods while supporting real-time deployment across the tested devices (Fig.~\ref{fig:teaser}).
In summary, our contributions are as follows:
\begin{itemize}
  \item We develop RTPolicy, an efficient end-to-end visuomotor policy that jointly reduces visual encoding, multi-view conditioning, and action-generation costs for precise manipulation.
  \item We introduce view-aware feature fusion and visual compression, where learnable view embeddings guide cross-view information exchange, and the fused representation is compressed before action generation.
  \item RTPolicy is evaluated in simulation and on a real bimanual platform, achieving the highest mean success rate among the compared methods while running in real time on the RTX 4090 and three edge devices.
\end{itemize}


\section{Related Work}

\subsection{Visuomotor Policies}

Visuomotor policies learn a direct mapping from visual observations and robot state to manipulation actions.
ACT~\cite{zhao2023learning} predicts action sequences with a Transformer and executes them in a receding horizon.
Diffusion Policy~\cite{chi2023diffusion} models action trajectories with conditional denoising and supports multimodal manipulation.
RVT~\cite{goyal2023rvt} constructs a three-dimensional representation from multiple views, while VO-DP~\cite{ni2025vodp} combines semantic DINOv2 features with geometric VGGT~\cite{wang2025vggt} features before diffusion control.

These methods improve temporal modeling and geometric reasoning, but they do not jointly address the high accuracy and low latency required for precise multi-view robotic manipulation.
Their visual backbones, geometric processing, or iterative action generation can increase deployment cost.
RTPolicy addresses this gap by reducing the visual conditioning path and the action generation cost while preserving the spatial information needed for precise manipulation.

\subsection{Efficient and Low Latency Manipulation Policies}

For visuomotor policies, one group of methods reduces the number of action generation steps.
Consistency Policy~\cite{prasad2024consistency} and One-Step Diffusion Policy~\cite{wang2024onedp} distill a diffusion trajectory into a consistency model or a single-step generator.
FlowPolicy~\cite{zhang2024flowpolicy} and ManiFlow~\cite{yan2025maniflow} learn consistency flow fields that support one or a few integration steps.
FreqPolicy~\cite{su2025freqpolicy} uses frequency consistency to preserve temporal structure during one-step generation.
Streaming Flow~\cite{jiang2025streamingflow} takes a different approach and emits actions during flow integration instead of waiting for the complete trajectory.
These methods reduce action generation latency, but they require distillation, consistency training, a modified action representation, or a specialized streaming solver.
Their visual encoder and multi-view feature fusion costs remain unchanged.

RTPolicy incorporates efficiency into the policy architecture.
The encoder performs view-aware feature fusion across views.
A visual compression module compresses the visual representation before flow-based action generation.
This joint design integrates efficiency into the visual conditioning and action generation paths for multi-view robotic manipulation.

\section{Method}

RTPolicy generates action chunks from multi-view images and observed robot states through a visual conditioning pipeline and a flow-based action model.
As illustrated in Fig.~\ref{fig:overview}, a frozen visual encoder extracts dense features, view-aware feature fusion exchanges information within and across views, and visual compression reduces the fused representation before action generation.
We first define the prediction task, then describe the visual representation and the training and inference procedures.

\subsection{Problem Formulation}

At control step $t$, the policy receives $V$ camera images $\{I_t^{(v)}\}_{v=1}^{V}$ and an observed robot-state prefix $S_t\in\mathbb{R}^{T_o\times d_a}$, where $T_o$ is the number of observed states.
In our joint-control setting, each state and action has $d_a$ dimensions.
The policy predicts an action chunk $A_t\in\mathbb{R}^{H_a\times d_a}$ with horizon $H_a$.
Throughout this section, $A_t$ contains only future actions. The observed prefix is a fixed condition rather than a prediction target.
The index $t$ denotes the control step, whereas $\tau\in[0,1]$ denotes the flow time used for action generation.

\subsection{Compact Multi-View Representation}

\subsubsection{View-Aware Feature Fusion}
A frozen DINOv3 ConvNeXt encoder~\cite{simeoni2025dinov3} maps each image to dense spatial tokens, $F_t^{(v)}=E(I_t^{(v)})\in\mathbb{R}^{P\times d_v}$, where $P$ is the number of tokens per view and $d_v$ is the feature width.
The encoder remains fixed during policy training, while the visual adaptation modules and action model learn from demonstrations.
To identify the source view, a learnable view embedding $e_v\in\mathbb{R}^{d_v}$ is added to every token of view $v$.
The features are then refined within each view and fused across views:
  \begin{equation}
  \begin{gathered}
  R_t^{(v)}
  = \mathcal{T}_{\mathrm{intra}}\!\left(F_t^{(v)}+e_v\right),\\
  \widetilde{F}_t
  = \mathcal{T}_{\mathrm{cross}}\!\left(
  \operatorname{Concat}_{v=1}^{V}R_t^{(v)}\right).
  \end{gathered}
  \label{eq:method_view_fusion}
  \end{equation}
$R_t^{(v)}\in\mathbb{R}^{P\times d_v}$ denotes the features refined within view $v$, while $\widetilde{F}_t\in\mathbb{R}^{VP\times d_v}$ denotes the concatenated feature sequence after cross-view fusion.
Here $\operatorname{Concat}$ joins sequences along the token dimension.
Both $\mathcal{T}_{\mathrm{intra}}$ and $\mathcal{T}_{\mathrm{cross}}$ consist of a single Transformer block with hidden width $d_v$ and $h$ attention heads.
The same $\mathcal{T}_{\mathrm{intra}}$ parameters are shared across all camera views and are applied independently to each view.
The cross-view module $\mathcal{T}_{\mathrm{cross}}$ then operates on the concatenated multi-view sequence, allowing each token to incorporate information from other views.
Thus, cross-view interaction occurs while the spatial representation is still dense, before visual compression reduces the number of visual tokens.

\subsubsection{Visual Compression}
The fused sequence retains the token ordering of the input views.
We extract the $P$ output tokens corresponding to each view as $\widetilde{F}_t^{(v)}$ and reshape them to that view's original spatial grid.
The visual compression module applies residual downsampling followed by adaptive average pooling, then flattens the pooled grid into $m$ tokens per view.
For each view, the compression module maps the fused spatial features to a compact token sequence,
\begin{equation}
\operatorname{Compress}
\left(\widetilde{F}_t^{(v)}\right)
\in
\mathbb{R}^{m \times d_c}.
\end{equation}
Here $\operatorname{Compress}$ denotes the visual compression mapping, and $d_c$ is the compressed feature width.
The compact visual condition is then formed as
\begin{equation}
C_t=
\operatorname{Concat}_{v=1}^{V}
\operatorname{Compress}
\left(\widetilde{F}_t^{(v)}\right)
\in
\mathbb{R}^{Vm \times d_c}.
\label{eq:method_scene_compression}
\end{equation}
The residual shortcut provides a direct feature path through the downsampling stage, while adaptive pooling controls the number of output tokens.
Although compression is applied separately to each view's grid, its inputs already contain cross-view information.
On the real platform, $V=3$, $P=300$, $m=4$, and $d_c=1024$, so the visual condition is reduced from $900$ dense tokens to $12$ compact visual tokens, i.e., $C_t\in\mathbb{R}^{12\times1024}$.
For three views, this reduces the visual condition from 900 to 12 tokens, a 75-fold reduction in token count.
The resulting compact visual condition $C_t$ is shared across all flow evaluations for the current observation.

\subsection{Flow-Based Action Generation}
\label{sec:action_generation}

\subsubsection{Conditioning the action transformer}
RTPolicy parameterizes the conditional velocity field with a Diffusion Transformer (DiT)~\cite{peebles2023dit}.
At flow time $\tau$, the network receives a noisy future action chunk $Z_\tau\in\mathbb{R}^{H_a\times d_a}$ together with the fixed state prefix $S_t$ and compact visual condition $C_t$.
The prefix and noisy actions are concatenated along the sequence dimension and projected into the DiT hidden space, with one token per state or action.
A sinusoidal encoding of $\tau$ is projected to a time token and prepended to this sequence, and positional embeddings encode the sequence order.
The compact visual tokens are projected to the same hidden width and supplied through cross-attention.
We write $v_\theta(Z_\tau,\tau;S_t,C_t)$ for the network output at the future action positions only, which has the same shape as $Z_\tau$.
The state prefix remains fixed during both training and inference.

\subsubsection{Flow matching objective}
We train the action model with flow matching~\cite{lipman2023flow} on normalized future action chunks.
For a demonstration chunk $A_t$ and Gaussian noise $\epsilon$ of the same shape, the interpolation path is
\begin{equation}
  Z_\tau=(1-\tau)\epsilon+\tau A_t,
  \qquad \epsilon\sim\mathcal{N}(0,I).
  \label{eq:method_flow_path}
\end{equation}
Here $I$ denotes the identity matrix with the same shape as the Gaussian noise.
Its target velocity is $A_t-\epsilon$.
The training objective supervises only the future action positions:
\begin{equation}
  \mathcal{L}_{\mathrm{FM}}=
  \mathbb{E}_{\mathrm{data},\tau,\epsilon}\!\left[
    \left\|v_\theta(Z_\tau,\tau;S_t,C_t)
    -(A_t-\epsilon)\right\|_F^2
  \right].
  \label{eq:method_flow_loss}
\end{equation}
The expectation is over demonstration samples, sampled flow times, and Gaussian noise; the squared Frobenius norm sums squared errors over future time steps and action dimensions.
The observed prefix is neither interpolated with noise nor included in the prediction loss.

\subsubsection{Four-step inference}
At inference, the future action slots are initialized with Gaussian noise, $Z^{(0)}\sim\mathcal{N}(0,I)$, while the observed state prefix remains fixed.
With $K$ uniform Euler steps, the future chunk is updated as
\begin{equation}
  \begin{aligned}
    Z^{(j+1)} &= Z^{(j)}+\frac{1}{K}
      v_\theta\!\left(Z^{(j)},\frac{j}{K};S_t,C_t\right),\\
    &\hspace{2em} j=0,\ldots,K-1.
  \end{aligned}
  \label{eq:method_flow_inference}
\end{equation}
Each network evaluation uses the same observed prefix and compact visual condition. Only the future action slots evolve. For implementation convenience, we form a full sequence by concatenating the fixed prefix and evolving future slots. The prefix positions are clamped to $S_t$ before every network evaluation, while only the velocity predictions at future positions are retained for Euler integration.
We use $K=4$, requiring four DiT evaluations per action chunk.
This budget balances action generation cost and task success rate, with the choice supported empirically by the integration-step ablation in Table~\ref{tab:nfe_ablation}.
After the final update, $Z^{(K)}$ is denormalized and passed to the controller as the predicted action chunk.

\section{Simulation Experiments}

\subsection{Experimental Setting}

\subsubsection{Benchmark and data}
We evaluate RTPolicy on RoboTwin~2.0~\cite{chen2025robotwin}, a benchmark for bimanual manipulation that contains 50 tasks with structured domain randomization.
We use the officially released demonstration split with 50 demonstrations for each target task.
Each observation contains three RGB images from the left, right, and head cameras at a resolution of $240\times320$, together with a 14-dimensional robot state.
The policy predicts the corresponding joint action trajectory.
We report the success rate over 100 evaluation episodes with the benchmark seed and average the success rates over tasks.

For comparison with the official single-task leaderboard, we use the clean2clean setting, where 50 clean demonstrations per task are used for training, and 100 trials are used for evaluation on Aloha AgileX.
The compared methods are $\pi_0$~\cite{black2024pi0}, RDT~\cite{liu2024rdt}, ACT~\cite{zhao2023learning}, and DP~\cite{chi2023diffusion}.
A matched comparison with DP and VO-DP~\cite{ni2025vodp} under the same training data is reported in the ablation studies.
In the tables below, BBH, AB, CAC, CB, DBB, and SO abbreviate beat block hammer, adjust bottle, click alarmclock, click bell, dump bin bigbin, and scan object, respectively.

\subsubsection{Training protocol}
\label{sec:training_protocol}
All reported task results use task-specific evaluation.
A checkpoint is trained for one target task and is evaluated only on that task.
Pretrained visual features provide broad coverage, while task-specific control still requires adaptation of the visual-to-action correspondence.
We therefore use joint pretraining as an initialization in the data-limited setting, followed by a separate fine-tuning stage for each target task.
In the full RoboTwin study, this uses the 50-task corpus before task-specific fine-tuning.

\subsubsection{Implementation details}
We train with AdamW and bfloat16 mixed precision.
Joint pretraining runs for 300K steps, and task-specific fine-tuning runs for 30K steps.
Joint pretraining uses a learning rate of $2\times10^{-4}$ and task fine-tuning uses $2\times10^{-5}$, with a batch size of 256.
The reported evaluation uses four flow integration steps.
The action model uses a 2048-dimensional DiT hidden space with eight blocks and 32 attention heads.
The flow time is sampled as $\tau=(0.999-s)/0.999$ with $s\sim\mathrm{Beta}(1.5,1.0)$, and the noise $\epsilon$ follows a standard Gaussian.
The simulation uses $H=30$ with $T_o=1$, and the real platform uses $H=46$ with $T_o=1$, so the predicted action chunk spans 45 steps at 30 fps, namely 1.5 seconds.
State and action values are normalized with statistics from the corresponding training protocol.

\begin{table}[t]
\caption{Clean2clean single-task comparison on RoboTwin~2.0.  Per task rows
show ten representative tasks.  The two separated blocks report the mean
over these ten tasks and the mean over the full 50-task benchmark.  All
baselines are from the official leaderboard.  RTPolicy attains the highest mean on both scales.}
\label{tab:robotwin_selected}
\centering
\setlength{\tabcolsep}{1pt}
\resizebox{\linewidth}{!}{
    \begin{tabular}{lrrrrr}
    \toprule
    Task & RTPolicy & $\pi_0$~\cite{black2024pi0} & RDT~\cite{liu2024rdt} & ACT~\cite{zhao2023learning} & DP~\cite{chi2023diffusion} \\
    \midrule
    Adjust Bottle & \textbf{97} & 90 & 81 & \textbf{97} & \textbf{97} \\
    Beat Block Hammer & 76 & 43 & \textbf{77} & 56 & 42 \\
    Blocks Ranking RGB & 5 & \textbf{19} & 3 & 1 & 0 \\
    Blocks Ranking Size & 3 & \textbf{7} & 0 & 0 & 1 \\
    Click Alarmclock & \textbf{76} & 63 & 61 & 32 & 61 \\
    Click Bell & 73 & 44 & \textbf{80} & 58 & 54 \\
    Dump Bin Bigbin & \textbf{86} & 83 & 64 & 68 & 49 \\
    Grab Roller & 97 & 96 & 74 & 94 & \textbf{98} \\
    Handover Block & \textbf{45} & \textbf{45} & \textbf{45} & 42 & 10 \\
    Handover Mic & 72 & \textbf{98} & 90 & 85 & 53 \\
    \midrule
    Mean (ten tasks above) & \textbf{63.0} & 58.8 & 57.5 & 53.3 & 46.5 \\
    \midrule
    Mean (full 50 tasks) & \textbf{52.3} & 46.4 & 34.5 & 29.7 & 28.0 \\
    \bottomrule
    \end{tabular}
}
\end{table}

\subsection{Main Performance}

Table~\ref{tab:robotwin_selected} reports the clean2clean single-task comparison with the leaderboard baselines.
RTPolicy attains the highest mean success rate both on the ten representative tasks (63.0) and over the full 50-task benchmark (52.3), against 58.8 and 46.4 for the strongest baseline $\pi_0$.
The full training configuration attains the highest mean, and the advantage concentrates on tasks that demand accurate localization.
On click alarmclock, RTPolicy reaches 76 against 63 for $\pi_0$ and 32 for ACT.
On beat block hammer, RTPolicy reaches 76 against 43 for $\pi_0$ and 42 for DP.
On dump bin bigbin, RTPolicy reaches 86 against 64 for RDT and 49 for DP.
Ranking tasks such as blocks ranking remain difficult for every method, where the success rates of all policies stay below 20.
The per-task comparison reports ten representative tasks, and the mean over the full 50-task benchmark is reported in the same table.

\subsection{Ablation Studies}

The ablations examine three factors that determine the deployment behavior of RTPolicy, namely the amount and source of supervision, the number of flow evaluations, and the structure of the visual condition.

\begin{table}[t]
\caption{Success rate on beat block hammer as the number of demonstrations
grows from 50 to 100, where each method trains with its native recipe on
the same demonstration split.}
\label{tab:data_scaling}
\centering
\setlength{\tabcolsep}{16pt}
\resizebox{\linewidth}{!}{
    \begin{tabular}{lrr}
    \toprule
    Method & 50 demos & 100 demos \\
    \midrule
    DP~\cite{chi2023diffusion} & 13 & 38 \\
    VO-DP~\cite{ni2025vodp} & 52 & 76 \\
    RTPolicy & \textbf{64} & \textbf{85} \\
    \bottomrule
    \end{tabular}
}
\end{table}

\subsubsection{Data scaling and initialization}
Table~\ref{tab:data_scaling} compares the data scaling behavior of RTPolicy with DP and VO-DP on beat block hammer, where each method uses its respective training settings on the same demonstration split.
All methods improve substantially as the number of demonstrations grows from 50 to 100, with gains of 21 to 25 points, showing that more data improves visual-action alignment.
RTPolicy starts 12 points above VO-DP and 51 points above DP at 50 demonstrations, and reaches 85 against 76 and 38 at 100 demonstrations.
Its lead narrows to 9 and 47 points, respectively, but remains substantial as all methods benefit from additional data.
\begin{table}[t]
\caption{Success rate (\%) as a function of flow integration steps, trained
from scratch with 100 demonstrations per task on the five task suite.  BBH
= beat block hammer, AB = adjust bottle, CAC = click alarmclock, CB =
click bell, DBB = dump bin bigbin.}
\label{tab:nfe_ablation}
\centering
\setlength{\tabcolsep}{8pt}
\resizebox{\linewidth}{!}{
    \begin{tabular}{lrrrrrr}
    \toprule
    Task & \multicolumn{6}{c}{Integration Steps} \\
    \cmidrule(lr){2-7}
     & 2 & 3 & 4 & 5 & 6 & 10 \\
    \midrule
    BBH & 75 & 83 & 85 & \textbf{87} & 83 & 85 \\
    AB & 98 & 99 & \textbf{100} & \textbf{100} & \textbf{100} & \textbf{100} \\
    CAC & \textbf{58} & 56 & 53 & 53 & 51 & 50 \\
    CB & \textbf{81} & 80 & 79 & 77 & 75 & 69 \\
    DBB & \textbf{94} & 87 & 90 & 93 & \textbf{94} & 89 \\
    \midrule
    Mean & 81.2 & 81.0 & 81.4 & \textbf{82.0} & 80.6 & 78.6 \\
    \bottomrule
    \end{tabular}
}
\end{table}

\subsubsection{Inference steps}
Flow matching allows the action trajectory to be integrated with different numbers of function evaluations.
The sweep in Table~\ref{tab:nfe_ablation} uses the same checkpoint, trained from scratch with 100 demonstrations per task, and changes only the integration budget.
From $K=2$ to $K=6$, the mean success rate stays within a $1.4$-point band ($80.6$--$82.0$) and remains above the ten-step default of $78.6$.
Four steps keep the mean within this band and use $40\%$ of the DiT evaluations of the ten-step scheduler, so $K=4$ is the working point supported by the ablation.

\begin{table}[t]
\caption{Ablation with 200 demonstrations per task.
Small, Base, and Large denote the corresponding DINOv3 ConvNeXt encoder scales, and the last column reports the encoder speedup relative to Large.
BBH = beat block hammer, CAC = click alarmclock, and SO = scan object.}
\label{tab:backbone_ablation}
\centering
\setlength{\tabcolsep}{4pt}
\resizebox{\linewidth}{!}{
    \begin{tabular}{lccccc}
    \toprule
    Backbone & BBH & CAC & SO & Mean & Speedup \\
    \midrule
    Small & 74 & 69 & \textbf{60} & 67.7 & \textbf{3.25$\times$} \\
    Base & 93 & 65 & 58 & 72.0 & 2.15$\times$ \\
    Large & \textbf{94} & \textbf{74} & 55 & \textbf{74.3} & 1.00$\times$ \\
    \bottomrule
    \end{tabular}
}
\end{table}

\subsubsection{Visual backbone}
Table~\ref{tab:backbone_ablation} compares the small, base, and large DINOv3 ConvNeXt encoders.
This comparison uses 200 demonstrations per task, where the demonstration data are no longer the limiting factor, so the scores reflect backbone capacity.
The large encoder reaches 74.3 against 72.0 for the base encoder, with a 2.15 times higher encoder latency.
The small encoder is weaker in accuracy.
We therefore use the base encoder in the main benchmark, which keeps most of the accuracy at a lower latency.

\begin{table}[t]
\caption{Ablation of view-aware feature fusion and visual compression in RTPolicy.
All variants are trained from scratch with 100 demonstrations per task and evaluated with four integration steps.
BBH = beat block hammer, AB = adjust bottle, CAC = click alarmclock, CB = click bell, DBB = dump bin bigbin.}
\label{tab:visual_ablation}
\centering
\setlength{\tabcolsep}{4pt}
\resizebox{\linewidth}{!}{
    \begin{tabular}{ccrrrrrr}
    \toprule
    \multirow{2}{*}{\shortstack{View-aware\\Feature Fusion}} & \multirow{2}{*}{\shortstack{Visual\\Compression}} &
    \multicolumn{5}{c}{Task} & \multirow{2}{*}{Mean} \\
    \cmidrule(lr){3-7}
    & & BBH & AB & CAC & CB & DBB & \\
    \midrule
    $\checkmark$ & $\checkmark$ & 85 & \textbf{100} & 53 & \textbf{79} & \textbf{90} & 81.4 \\
    $\checkmark$ & $\times$ & 84 & \textbf{100} & \textbf{83} & 67 & 88 & \textbf{84.4} \\
    $\times$ & $\checkmark$ & 83 & 99 & 64 & 55 & 89 & 78.0 \\
    $\times$ & $\times$ & \textbf{89} & 99 & 49 & 54 & 86 & 75.4 \\
    \bottomrule
    \end{tabular}
}
\end{table}

\begin{figure}[t]
  \centering
  \begin{subfigure}[t]{\columnwidth}
    \centering
    \includegraphics[width=\columnwidth]{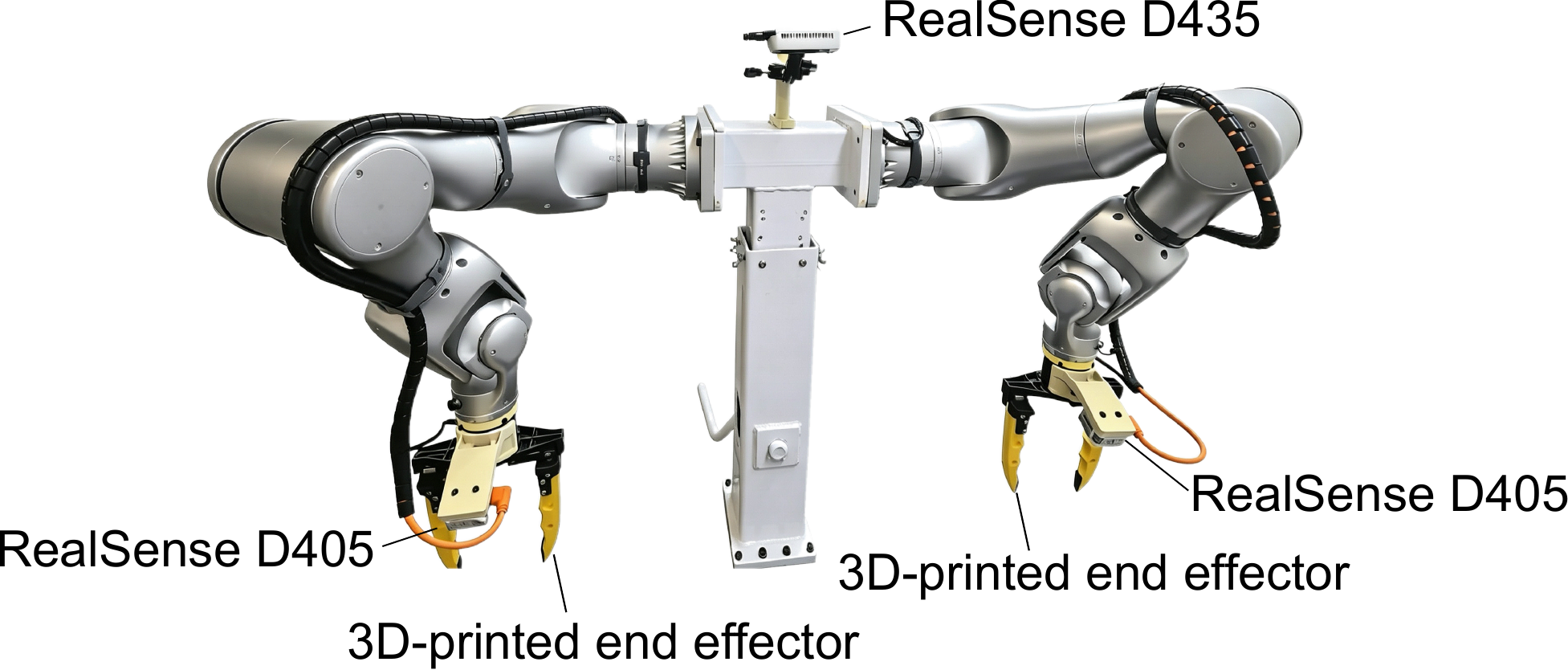}
    \caption{Bimanual robot platform.}
    \label{fig:real_platform}
  \end{subfigure}
  \vspace{0.4em}
  \begin{subfigure}[t]{\columnwidth}
    \centering
    \includegraphics[width=\columnwidth]{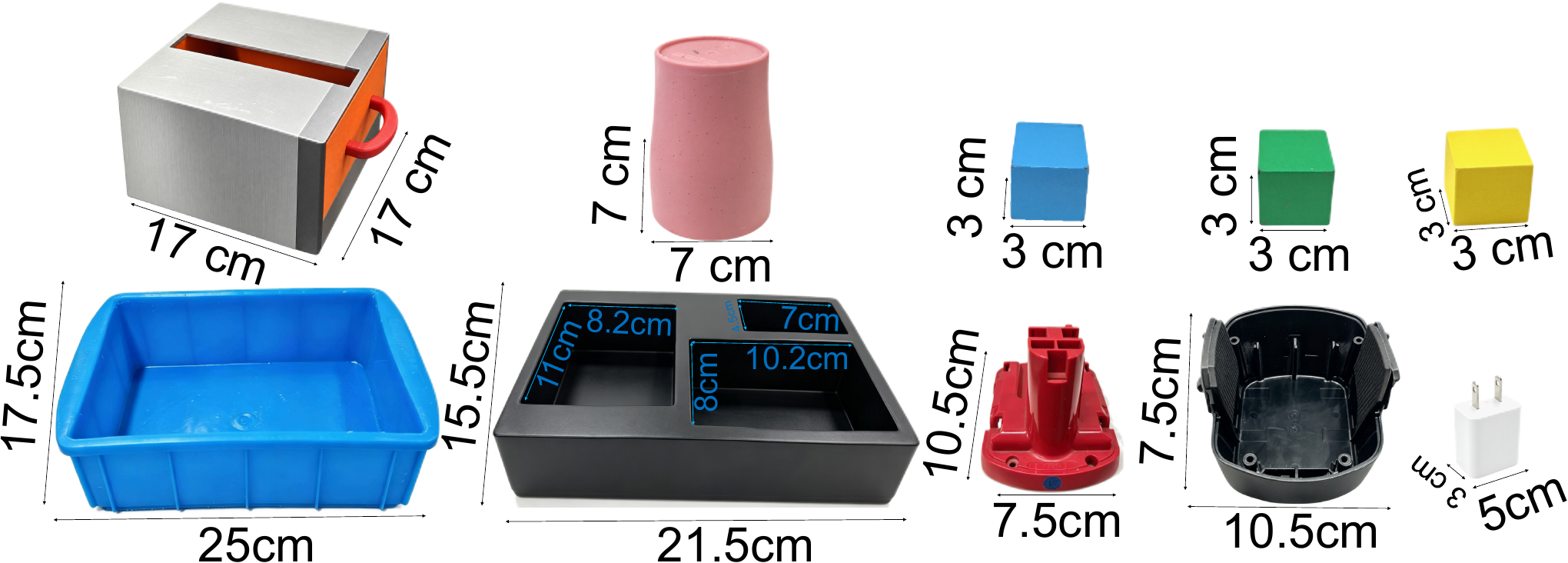}
    \caption{Objects used in the real tasks.}
    \label{fig:real_objects}
  \end{subfigure}
  \caption{Real-world hardware and objects.
  The platform is equipped with two ROKAE AR5 arms and three cameras.
  A head camera is mounted above the workspace, and a wrist camera is attached to each gripper.
  The objects of the three evaluation tasks are shown with their physical sizes.}
  \vspace{-4mm}
  \label{fig:real_setup}
\end{figure}

\begin{figure*}[t]
  \centering
  \includegraphics[width=\textwidth]{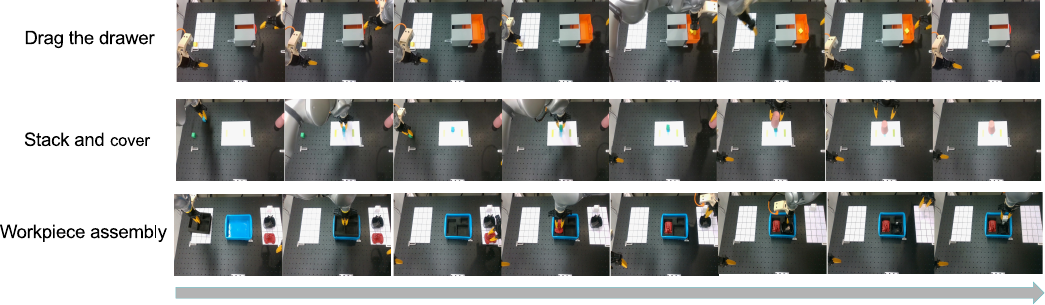}
  \caption{Real-world tasks.
  The rows show drag the drawer, stack and cover, and workpiece assembly from top to bottom.}
  \vspace{-2mm}
  \label{fig:real_tasks}
\end{figure*}

\subsubsection{View-aware feature fusion and visual compression}
\label{sec:visual_ablation}
Table~\ref{tab:visual_ablation} reports the component ablation, which removes the view-aware feature fusion module, the visual compression module, or both.
View-aware feature fusion is the dominant component.
It alone lifts click alarmclock from 49 to 83, and keeping feature fusion without visual compression gives the best mean of 84.4.
Once feature fusion is present, visual compression changes the mean only from 84.4 to 81.4, whereas without feature fusion the model drops to 78.0, and removing both modules reaches 75.4, so at least one visual adaptation module is required.
In this simulation setting, the data are small, and the fused tokens are cheap to condition on, so the compression benefit of the visual compression module is limited.
On real deployments with high-resolution inputs, the visual compression module compresses the fused tokens and suppresses redundant background evidence before the action model, which reduces the conditioning cost and inference time.
RTPolicy therefore combines view-aware feature fusion with visual compression.

\section{Real World Experiments}
\label{sec:real_experiments}

\subsection{Experimental Setting}

\subsubsection{Platform and tasks}
We deploy RTPolicy on a bimanual platform with two ROKAE AR5 arms, each with seven degrees of freedom and a parallel-jaw gripper, as shown in Fig.~\ref{fig:real_platform}.
Three RGB cameras, one head camera and two wrist cameras, capture $480\times640$ images at 30 fps.
The robot exposes a state vector with 16 dimensions, and the policy predicts a joint action vector of the same dimension.
For each task, we collect 200 teleoperated demonstrations and train a task-specific checkpoint directly on the real data.

We evaluate on three tasks, as shown in Fig.~\ref{fig:real_tasks}.
Drag the drawer requires the right arm to pull the drawer open and retreat, the left arm to grasp a block and place it inside, and the right arm to push the drawer closed, so the task stresses sequential bimanual coordination across three stages.
Stack and cover requires stacking two blocks and covering them with a cup, where the placement tolerance is small, and the stack is easily toppled.
Workpiece assembly requires the left arm to place a mat inside a base and the right arm to insert components into slots, which is a long-horizon task and demands fine placement.
The objects used in these tasks are shown in Fig.~\ref{fig:real_objects}.
Each task is evaluated with 50 trials on the real robot, and a trial is successful only if all stages of the task are completed.

\subsubsection{Compared methods}
All compared methods are trained on the same 200 demonstrations of each task and evaluated under the same setting.
DP~\cite{chi2023diffusion} and VO-DP~\cite{ni2025vodp} are visuomotor policies that share the matched single-task setting.
They support a single camera view only, whereas RTPolicy and $\pi_{0.5}$ consume three views, so the comparison reports each method in its native camera configuration.
$\pi_{0.5}$~\cite{black2025pi05} is fine-tuned on the same demonstrations as a single-task checkpoint.
RTPolicy is trained for 150K steps on each task.

\begin{table}[t]
\caption{Real-world success rate comparison on the bimanual ROKAE AR5.
All methods are trained on the same 200 demonstrations per task
with 150K steps and evaluated with 50 trials per task on the RTX 4090.}
\label{tab:real_comparison}
\centering
\setlength{\tabcolsep}{2pt}
\resizebox{\linewidth}{!}{
    \begin{tabular}{lcccc}
    \toprule
    Method & Drag Drawer $\uparrow$ & Stack and Cover $\uparrow$ &
    Workpiece Assembly $\uparrow$ & Mean $\uparrow$ \\
    \midrule
    DP~\cite{chi2023diffusion} 
      & 80 & 8 & 0 & 29.3 \\
    VO-DP~\cite{ni2025vodp} 
      & 84 & 34 & 4 & 40.7 \\
    $\pi_{0.5}$~\cite{black2025pi05} 
      & 88 & 56 & 12 & 52.0 \\
    RTPolicy 
      & \textbf{92} & \textbf{94} & \textbf{30} & \textbf{72.0} \\
    \bottomrule
    \end{tabular}
}
\end{table}

\begin{table*}[t]
\caption{Policy latency (ms) and frame rate (Hz, in parentheses) across
four deployment platforms.  Views is the number of camera views consumed by
each method, and Tokens is the number of visual tokens that enter action
generation.  GPU platforms run TensorRT engines at FP32 and
FP16, and BPU platforms run quantized INT8 and INT16 models.  A dash
indicates that running on the board is too slow to be practical for
deployment.}
\label{tab:efficiency_all}
\centering
\setlength{\tabcolsep}{3pt}
\resizebox{\linewidth}{!}{
    \begin{tabular}{lcccccccccc}
    \toprule
    \multirow{2}{*}{Method} &
    \multirow{2}{*}{Views} &
    \multirow{2}{*}{Tokens} &
    \multicolumn{2}{c}{RTX 4090} &
    \multicolumn{2}{c}{Jetson Orin NX 16GB} &
    \multicolumn{2}{c}{RDK S600} &
    \multicolumn{2}{c}{RDK S100} \\
    \cmidrule(lr){4-5} \cmidrule(lr){6-7} \cmidrule(lr){8-9} \cmidrule(lr){10-11}
     & & & FP32 & FP16 & FP32 & FP16 & INT8 & INT16 & INT8 & INT16 \\
    \midrule
    DP~\cite{chi2023diffusion}
      & 1 & 1
      & 23.15 (43.2) & 8.72 (114.7)
      & 49.16 (20.3) & 31.25 (32.0)
      & 16.40 (61.0) & 19.76 (50.6)
      & 67.23 (14.9) & 77.98 (12.8) \\
    VO-DP~\cite{ni2025vodp}
      & 1 & 1
      & 88.16 (11.3) & 25.31 (39.5)
      & -- & 410.34 (2.4)
      & 79.97 (12.5) & 181.47 (5.5)
      & 489.0 (2.0) & 1025.38 (1.0) \\
    $\pi_{0.5}$~\cite{black2025pi05}
      & 3 & 768
      & 109.33 (9.15) & 58.84 (17.0)
      & -- & --
      & 135.5 (7.4) & 331.6 (3.0)
      & 778.10 (1.3) & 1491.59 (0.7) \\
    RTPolicy
      & 3 & 12
      & 23.80 (42.0) & 10.92 (91.6)
      & 266.3 (3.8) & 143.2 (7.0)
      & 18.59 (53.8) & 30.35 (32.9)
      & 86.56 (11.6) & 149.72 (6.7) \\
    \bottomrule
    \end{tabular}
    \vspace{-2mm}
}
\end{table*}

\subsubsection{Streaming deployment}
During deployment, the policy runs in a streaming mode.
Inference proceeds at a fixed rate of 10 Hz on the latest observation, in parallel with execution at 30 Hz, so a new action chunk is generated while the robot is still executing the current one.
Each chunk is anchored to the wall-clock time of its observation.
A stitching module then replans a smooth transition from the currently executed trajectory to the new chunk.
The stitched trajectory is obtained by solving, per joint channel, a piecewise jerk optimization~\cite{werling2010optimal} formulated as
\begin{equation}
\min_{\{x_k\}}
\sum_{k} \left(x_k-r_k\right)^{2}
+w_a\sum_{k}\left(\frac{\ddot{x}_k}{a_{\max}}\right)^{2}
+w_j\sum_{k}\left(\frac{u_k}{j_{\max}}\right)^{2},
\label{eq:stitching}
\end{equation}
where $x_k$ denotes the joint position at knot $k$ of the stitched trajectory sampled on a uniform grid, $r_k$ is the new chunk linearly interpolated at the same knot time, $\ddot{x}_k$ and $u_k$ are the acceleration and the piecewise constant jerk of the stitched trajectory, and $w_a$, $w_j$, $a_{\max}$, and $j_{\max}$ are the smoothing weights and limits.
The boundary state at the switching time is constrained to match the position, velocity, and acceleration of the active trajectory, so the output is $C^2$ continuous at the junction.
The acceleration and jerk limits are enforced as box constraints in the optimizer in addition to their normalization role in the cost, and the gripper channels use separate limits from the rotational joints.
With this design, the delay from a new observation to a corrective action is one inference cycle rather than the duration of a whole chunk, and errors accumulated within a chunk are corrected by the next chunk conditioned on fresh observations.

\subsection{Results}
Table~\ref{tab:real_comparison} reports the success rate of all methods on the three tasks.
RTPolicy attains the highest score on every task and the best mean of 72.0.
The gap to the baselines widens as the demand for precise visual correction grows.
On workpiece assembly, the most demanding task with small placement tolerance and a long stage sequence, RTPolicy reaches 30 against 12 for $\pi_{0.5}$ and 4 for VO-DP, while DP fails all trials.
On stack and cover, RTPolicy reaches 94 against 56 for $\pi_{0.5}$, 34 for VO-DP, and 8 for DP.
Using the same RTPolicy model on stack and cover, we vary the asynchronous update rate while keeping the task and model fixed.
At 10 Hz, inference finishes before the 100 ms update period and waits for the next scheduled update, while lower rates lengthen the interval between fresh observations.
The success rates are 94\%, 88\%, and 70\% at 10, 5, and 2 Hz, respectively.
This result shows that timely closed-loop correction is necessary and highlights the benefit of low inference latency.
The streaming deployment is what makes these results reachable in practice.
The three tasks are all multi-stage, and a single chunk is too long to bridge the inter-stage transitions, so a policy that replans only after finishing a chunk cannot correct its plan when the previous stage ends in a slightly displaced configuration.
With the streaming mode, each new chunk is conditioned on a fresh observation within one inference cycle, so the displacement is corrected before it propagates into the next stage.
Lower inference latency further shortens this correction cycle provided by RTPolicy.
$\pi_{0.5}$ attains a higher mean than DP and VO-DP, but its high inference latency delays the visual correction, which hurts most on the multi-stage tasks where the correction must arrive before the next stage begins.

We also evaluate the compression-free variant on stack and cover, which reaches 40 against 94 for the full model.
Moreover, its inference latency on the RTX 4090 grows by about 14 percent.
In the simulation study of Sec.~\ref{sec:visual_ablation}, visual compression appears dispensable at $240\times320$ inputs, whereas with $480\times640$ inputs the high-resolution tokens overload the visual condition without compression, so visual compression becomes necessary on the real platform.
This confirms the deployment rationale for keeping visual compression in RTPolicy.


\section{Efficiency}
\label{sec:efficiency}

We benchmark the end-to-end policy latency of all four methods on four deployment platforms, namely the NVIDIA RTX 4090 GPU, the Jetson Orin NX 16GB board, and the RDK S600 and RDK S100 BPU boards.
All measurements use a batch size of 1 and report the single-frame closed-loop latency, namely the time from camera observations to the action chunk.
The latency of each method is measured with the camera configuration of its real-world deployment in Sec.~\ref{sec:real_experiments}.
DP and VO-DP consume a single $480\times640$ camera, while RTPolicy and $\pi_{0.5}$ consume three $480\times640$ cameras.
Before action generation, DP and VO-DP compress the visual features into a single conditioning vector through global average pooling, $\pi_{0.5}$ keeps a 768-token visual sequence in the action model, and RTPolicy compresses its three views into 12 compact tokens.
For DP and VO-DP, the inference-time denoising process is replaced with a ten-step DDIM sampler~\cite{song2021ddim}.
RTPolicy and $\pi_{0.5}$ generate action chunks with the flow integration steps used in their real-world deployment.
GPU platforms run TensorRT~\cite{nvidia_tensorrt} engines at FP32 and FP32, and BPU platforms run quantized INT8 and INT16 models.
The FP32 column follows the default TensorRT execution mode. 
Table~\ref{tab:efficiency_all} reports the latency of all methods on the four platforms.

\paragraph{RTX 4090}
RTPolicy runs at 10.92 ms in FP32, which is on par with DP and 2.3 times faster than VO-DP and 5.4 times faster than $\pi_{0.5}$.
A frame rate of 91.6 Hz far exceeds the inference rate required by the streaming deployment, so the full pipeline is computation-bound only on the observation side.

\paragraph{Jetson Orin NX}
The Jetson Orin NX 16GB board is a 25 W class embedded platform with 16 GB of unified memory.
RTPolicy runs at 143.2 ms in FP32, which supports a 7 Hz inference rate.
The correction cycle is longer than at the 10 Hz rate, but still covers general manipulation frequencies.
VO-DP drops to 2.4 Hz in FP32 and $\pi_{0.5}$ runs too slowly on the board to be practical for deployment, so only DP and RTPolicy remain deployable at this power budget.

\paragraph{RDK S600}
The RDK S600 board carries a 4-core BPU that delivers 560 TOPS and runs quantized models natively.
DP runs fastest at 16.40 ms in INT8.
RTPolicy follows at 18.59 ms, which leaves a large margin over both the 10 Hz streaming inference rate and the 30 Hz execution rate.
VO-DP sustains the rate in INT8 at 12.5 Hz but drops to 5.5 Hz in INT16, and $\pi_{0.5}$ runs at 7.4 Hz in INT8 and 3.0 Hz in INT16, so both correct the accumulated deviations more slowly at their usable precisions, since their backbones dominate the compute cost.

\paragraph{RDK S100}
The RDK S100 board is an entry-level platform with an 80 TOPS BPU.
RTPolicy still runs at 86.56 ms in INT8 and keeps the 10 Hz streaming inference rate, while VO-DP and $\pi_{0.5}$ run at about 2 Hz or below and no longer provide timely correction.
Across the four platforms, RTPolicy sustains the 10 Hz streaming inference rate in FP32 on the GPU platforms and in INT8 on the BPU platforms, and still runs at 7 Hz on the Jetson board.
The success rate is measured with the GPU configuration, where RTPolicy attains the highest mean among the compared methods.

\section{Conclusion}
\label{sec:conclusion}

We presented RTPolicy, a visuomotor policy that combines a frozen DINOv3 ConvNeXt backbone, view-aware feature fusion, and visual compression with a flow-matching DiT action head to generate action chunks in four integration steps.
RTPolicy attains the highest mean success rate among the compared methods, 63.0 on the ten representative RoboTwin 2.0 tasks and 72.0 on the real bimanual platform.
The compact visual condition and four-step integration reduce inference latency and support timely correction during streaming deployment.
RTPolicy meets the 10 Hz inference budget on the RTX 4090, RDK S600, and RDK S100, and reaches 7 Hz on the Jetson Orin NX.
Together, these results show that RTPolicy combines high manipulation accuracy with low-latency action generation.

\textbf{{Limitations.}}
RTPolicy uses a frozen visual representation from self-supervised pretraining, while its action model is learned from task-specific demonstrations.
This training setup remains task-specific and does not provide a single task-conditioned generalist checkpoint.
Extending RTPolicy to generalist control will require pretraining the action model on large-scale multi-robot and real-world corpora together with a task-conditioning signal.






\bibliographystyle{IEEEtran}
\bibliography{refs}

@inproceedings{ross2010efficient,
  title={Efficient Reductions for Imitation Learning},
  author={Ross, Stephane and Bagnell, Drew},
  booktitle={Proceedings of the Thirteenth International Conference on Artificial Intelligence and Statistics},
  pages={661--668},
  volume={9},
  series={Proceedings of Machine Learning Research},
  publisher={PMLR},
  year={2010}
}

@inproceedings{ross2011reduction,
  title={A Reduction of Imitation Learning and Structured Prediction to No-Regret Online Learning},
  author={Ross, St{\'e}phane and Gordon, Geoffrey J. and Bagnell, J. Andrew},
  booktitle={Proceedings of the Fourteenth International Conference on Artificial Intelligence and Statistics},
  pages={627--635},
  volume={15},
  series={Proceedings of Machine Learning Research},
  publisher={PMLR},
  year={2011}
}

@inproceedings{zhao2023learning,
  title={Learning Fine-Grained Bimanual Manipulation with Low-Cost Hardware},
  author={Zhao, Tony Z. and Kumar, Vikash and Levine, Sergey and Finn, Chelsea},
  booktitle={Robotics: Science and Systems},
  year={2023},
  doi={10.15607/RSS.2023.XIX.016}
}

@inproceedings{chi2023diffusion,
  title={Diffusion Policy: Visuomotor Policy Learning via Action Diffusion},
  author={Chi, Cheng and Feng, Siyuan and Du, Yilun and Xu, Zhenjia and Cousineau, Eric and Burchfiel, Benjamin and Song, Shuran},
  booktitle={Robotics: Science and Systems},
  year={2023},
  doi={10.15607/RSS.2023.XIX.026}
}

@inproceedings{black2024pi0,
  title={$\pi_0$: A Vision-Language-Action Flow Model for General Robot Control},
  author={Black, Kevin and Brown, Noah and Driess, Danny and Esmail, Adnan and Equi, Michael Robert and Finn, Chelsea and Fusai, Niccolo and Groom, Lachy and Hausman, Karol and Ichter, Brian and Jakubczak, Szymon and Jones, Tim and Ke, Liyiming and Levine, Sergey and Li-Bell, Adrian and Mothukuri, Mohith and Nair, Suraj and Pertsch, Karl and Shi, Lucy Xiaoyang and Smith, Laura and Tanner, James and Vuong, Quan and Walling, Anna and Wang, Haohuan and Zhilinsky, Ury},
  booktitle={Robotics: Science and Systems},
  year={2025},
  doi={10.15607/RSS.2025.XXI.010}
}

@inproceedings{goyal2023rvt,
  title={{RVT}: Robotic View Transformer for {3D} Object Manipulation},
  author={Goyal, Ankit and Xu, Jie and Guo, Yijie and Blukis, Valts and Chao, Yu-Wei and Fox, Dieter},
  booktitle={Proceedings of The 7th Conference on Robot Learning},
  pages={694--710},
  volume={229},
  series={Proceedings of Machine Learning Research},
  publisher={PMLR},
  year={2023}
}

@inproceedings{liu2024rdt,
  title={{RDT-1B}: a Diffusion Foundation Model for Bimanual Manipulation},
  author={Liu, Songming and Wu, Lingxuan and Li, Bangguo and Tan, Hengkai and Chen, Huayu and Wang, Zhengyi and Xu, Ke and Su, Hang and Zhu, Jun},
  booktitle={International Conference on Learning Representations},
  year={2025}
}

@inproceedings{prasad2024consistency,
  title={Consistency Policy: Accelerated Visuomotor Policies via Consistency Distillation},
  author={Prasad, Aaditya and Lin, Kevin and Wu, Jimmy and Zhou, Linqi and Bohg, Jeannette},
  booktitle={Robotics: Science and Systems},
  year={2024},
  doi={10.15607/RSS.2024.XX.071}
}

@inproceedings{yan2025maniflow,
  title={{ManiFlow}: A General Robot Manipulation Policy via Consistency Flow Training},
  author={Yan, Ge and Zhu, Jiyue and Deng, Yuquan and Yang, Shiqi and Qiu, Ri-Zhao and Cheng, Xuxin and Memmel, Marius and Krishna, Ranjay and Goyal, Ankit and Wang, Xiaolong and Fox, Dieter},
  booktitle={Proceedings of The 9th Conference on Robot Learning},
  pages={2268--2293},
  volume={305},
  series={Proceedings of Machine Learning Research},
  publisher={PMLR},
  year={2025}
}

@inproceedings{wang2025vggt,
  title={{VGGT}: Visual Geometry Grounded Transformer},
  author={Wang, Jianyuan and Chen, Minghao and Karaev, Nikita and Vedaldi, Andrea and Rupprecht, Christian and Novotny, David},
  booktitle={IEEE/CVF Conference on Computer Vision and Pattern Recognition},
  year={2025}
}

@article{ni2025vodp,
  title={{VO-DP}: Semantic-Geometric Adaptive Diffusion Policy for Vision-Only Robotic Manipulation},
  author={Ni, Zehao and He, Yonghao and Qian, Lingfeng and Mao, Jilei and Fu, Fa and Sui, Wei and Su, Hu and Peng, Junran and Wang, Zhipeng and He, Bin},
  journal={arXiv preprint arXiv:2510.15530},
  year={2025}
}

@article{simeoni2025dinov3,
  title={{DINOv3}},
  author={Sim{\'e}oni, Oriane and Vo, Huy V. and Seitzer, Maximilian and Baldassarre, Federico and Oquab, Maxime and Jose, Cijo and Khalidov, Vasil and Szafraniec, Marc and Yi, Seungeun and Ramamonjisoa, Micha{\"e}l and Massa, Francisco and Haziza, Daniel and Wehrstedt, Luca and Wang, Jianyuan and Darcet, Timoth{\'e}e and Moutakanni, Th{\'e}o and Sentana, Leonel and Roberts, Claire and Vedaldi, Andrea and Tolan, Jamie and Brandt, John and Couprie, Camille and Mairal, Julien and J{\'e}gou, Herv{\'e} and Labatut, Patrick and Bojanowski, Piotr},
  journal={Transactions on Machine Learning Research},
  year={2026}
}

@inproceedings{chen2025robotwin,
  title={{RoboTwin} 2.0: A Scalable Data Generator and Benchmark with Strong Domain Randomization for Robust Bimanual Robotic Manipulation},
  author={Chen, Tianxing and Chen, Zanxin and Chen, Baijun and Cai, Zijian and Liu, Yibin and Li, Zixuan and Liang, Qiwei and Lin, Xianliang and Ge, Yiheng and Gu, Zhenyu and Deng, Weiliang and Guo, Yubin and Nian, Tian and Xie, Xuanbing and Chen, Qiangyu and Su, Kailun and Xu, Tianling and Liu, Guodong and Hu, Mengkang and Gao, Huan-ang and Wang, Kaixuan and Liang, Zhixuan and Qin, Yusen and Yang, Xiaokang and Luo, Ping and Mu, Yao},
  booktitle={International Conference on Machine Learning},
  year={2026}
}

@inproceedings{lipman2023flow,
  title={Flow Matching for Generative Modeling},
  author={Lipman, Yaron and Chen, Ricky T. Q. and Ben-Hamu, Heli and Nickel, Maximilian and Le, Matt},
  booktitle={International Conference on Learning Representations},
  year={2023}
}

@inproceedings{jiang2025streamingflow,
  title={Streaming Flow Policy: Simplifying Diffusion/Flow-Matching Policies by Treating Action Trajectories as Flow Trajectories},
  author={Jiang, Sunshine and Fang, Xiaolin and Roy, Nicholas and Lozano-P{\'e}rez, Tom{\'a}s and Kaelbling, Leslie Pack and Ancha, Siddharth},
  booktitle={Proceedings of The 9th Conference on Robot Learning},
  pages={238--257},
  volume={305},
  series={Proceedings of Machine Learning Research},
  publisher={PMLR},
  year={2025}
}

@inproceedings{su2025freqpolicy,
  title={{FreqPolicy}: Efficient Flow-based Visuomotor Policy via Frequency Consistency},
  author={Su, Yifei and Liu, Ning and Chen, Dong and Zhao, Zhen and Wu, Kun and Li, Meng and Xu, Zhiyuan and Che, Zhengping and Tang, Jian},
  booktitle={Advances in Neural Information Processing Systems},
  year={2025}
}

@inproceedings{wang2024onedp,
  title={One-Step Diffusion Policy: Fast Visuomotor Policies via Diffusion Distillation},
  author={Wang, Zhendong and Li, Max and Mandlekar, Ajay and Xu, Zhenjia and Fan, Jiaojiao and Narang, Yashraj and Fan, Linxi and Zhu, Yuke and Balaji, Yogesh and Zhou, Mingyuan and Liu, Ming-Yu and Zeng, Yu},
  booktitle={Proceedings of the 42nd International Conference on Machine Learning},
  pages={63399--63416},
  volume={267},
  series={Proceedings of Machine Learning Research},
  publisher={PMLR},
  year={2025}
}

@inproceedings{zhang2024flowpolicy,
  title={{FlowPolicy}: Enabling Fast and Robust {3D} Flow-based Policy via Consistency Flow Matching for Robot Manipulation},
  author={Zhang, Qinglun and Liu, Zhen and Fan, Haoqiang and Liu, Guanghui and Zeng, Bing and Liu, Shuaicheng},
  booktitle={Proceedings of the AAAI Conference on Artificial Intelligence},
  volume={39},
  number={14},
  pages={14754--14762},
  year={2025},
  doi={10.1609/aaai.v39i14.33617}
}

@inproceedings{peebles2023dit,
  title={Scalable Diffusion Models with Transformers},
  author={Peebles, William and Xie, Saining},
  booktitle={Proceedings of the IEEE/CVF International Conference on Computer Vision},
  pages={4195--4205},
  year={2023},
  doi={10.1109/ICCV51070.2023.00387}
}

@inproceedings{black2025pi05,
  title={$\pi_{0.5}$: a Vision-Language-Action Model with Open-World Generalization},
  author={Black, Kevin and Brown, Noah and Darpinian, James and Dhabalia, Karan and Driess, Danny and Esmail, Adnan and Equi, Michael Robert and Finn, Chelsea and Fusai, Niccolo and Galliker, Manuel Y. and Ghosh, Dibya and Groom, Lachy and Hausman, Karol and Ichter, Brian and Jakubczak, Szymon and Jones, Tim and Ke, Liyiming and LeBlanc, Devin and Levine, Sergey and Li-Bell, Adrian and Mothukuri, Mohith and Nair, Suraj and Pertsch, Karl and Ren, Allen Z. and Shi, Lucy Xiaoyang and Smith, Laura and Springenberg, Jost Tobias and Stachowicz, Kyle and Tanner, James and Vuong, Quan and Walke, Homer and Walling, Anna and Wang, Haohuan and Yu, Lili and Zhilinsky, Ury},
  booktitle={Proceedings of The 9th Conference on Robot Learning},
  pages={17--40},
  volume={305},
  series={Proceedings of Machine Learning Research},
  publisher={PMLR},
  year={2025}
}

@inproceedings{werling2010optimal,
  title={Optimal Trajectory Generation for Dynamic Street Scenarios in a Fren{\'e}t Frame},
  author={Werling, Moritz and Ziegler, Julius and Kammel, S{\"o}ren and Thrun, Sebastian},
  booktitle={IEEE International Conference on Robotics and Automation},
  pages={987--993},
  year={2010},
  doi={10.1109/ROBOT.2010.5509799}
}

@misc{nvidia_tensorrt,
  author       = {{NVIDIA Corporation}},
  title        = {{NVIDIA TensorRT}},
  year         = {2024},
  howpublished = {\url{https://developer.nvidia.com/tensorrt}},
  note         = {Version 10.3, accessed 2026-09-12}
}

@inproceedings{song2021ddim,
  author    = {Song, Jiaming and Meng, Chenlin and Ermon, Stefano},
  title     = {Denoising Diffusion Implicit Models},
  booktitle = {International Conference on Learning Representations},
  year      = {2021}
}

\end{document}